\documentclass{article}
\usepackage{ijcai24}

\usepackage{times}
\usepackage{soul}
\usepackage{url}
\usepackage[hidelinks]{hyperref}
\usepackage[utf8]{inputenc}
\usepackage[small]{caption}
\usepackage{graphicx}
\usepackage{amsmath}
\usepackage{amsthm}
\usepackage{booktabs}
\usepackage{algorithm}
\usepackage{algorithmic}
\usepackage[switch]{lineno}
\usepackage{subcaption}

\title{Choosing a Classical Planner with Graph Neural Networks}

\author{
Jana Vatter$^1$
\and
Ruben Mayer$^2$
\and
Hans-Arno Jacobsen$^3$
\and
Horst Samulowitz$^4$
\and
Michael Katz$^4$
\affiliations
$^1$Technical University of Munich\\
$^2$University of Bayreuth\\
$^3$University of Toronto\\
$^4$IBM Research
\emails
jana.vatter@tum.de,
ruben.mayer@uni-bayreuth.de,
jacobsen@eecg.toronto.edu,
samulowitz@us.ibm.com,
michael.katz1@ibm.com
}

\begin{document}

\maketitle

\begin{abstract}
    Online planner selection is the task of choosing a solver out of a predefined set for a given planning problem. As planning is computationally hard, the performance of solvers varies greatly on planning problems. Thus, the ability to predict their performance on a given problem is of great importance. While a variety of learning methods have been employed, for classical cost-optimal planning the prevailing approach uses Graph Neural Networks (GNNs). 
    In this work, we continue the line of work on using GNNs for online planner selection. We perform a  thorough investigation of the impact of the chosen GNN model, graph representation and node features, as well as prediction task.
    Going further, we propose using the graph representation obtained by a GNN as an input to the Extreme Gradient Boosting (XGBoost) model, resulting in a more resource-efficient yet accurate approach. We show the effectiveness of a variety of GNN-based online planner selection methods, opening up new exciting avenues for research on online planner selection.
\end{abstract}

\section{Introduction} 
Automated planning is a foundational discipline within the field of Artificial Intelligence (AI) research \cite{norvig2002modern}. The focus of planning lies in formulating goal-oriented policies, or, in the deterministic case, sequences of actions to achieve predefined goals. Applications range from robotics,  autonomous vehicles and industrial automation to incremental topology transformation \cite{ghallab2004automated,yoon2016planning,salehi2017incremental,rogalla2018improved,karpas2020automated,chakraborti2020robotic}.

Given the inherent complexity of classical planning, as described by Bylander et al. \cite{bylander1994computational}, there is unlikely to be a single planning algorithm that can work well across diverse planning problems. Over the years, a variety of planners were developed, tackling various aspects that make planning problems challenging. 
Consequently, there is a growing interest in the development of portfolio-based approaches \cite{seipp2012learning,vallati2012guide,cenamor2013learning,seipp2015automatic} where multiple planners are aggregated and out of this portfolio, a single planner or a schedule of planners are selected. Portfolio-based planning can be divided into \textit{offline} and \textit{online} methods. While \textit{offline} methods \cite{helmert2011fast,seipp2012learning} construct a single invocation schedule ahead of time which is anticipated to perform well across many domains, \textit{online} methods \cite{katz2018delfi,Ma2020,ferber2022explainable} adapt by learning to select the most suitable planner for each specific task. 

Focusing our attention on the online methods, a variety of 
deep learning based approaches were recently developed. There are Convolutional Neural Networks (CNNs) based approaches \cite{katz2018delfi,Sievers2019}, with the planner {\em Delfi} winning the cost-optimal track of the International Planning Competition (IPC) 2018, where each planning problem is represented by an image. 
However, these images are obtained by first converting the planning problem to a graph, for instance as a problem description graph (PDG) \cite{pochter2011exploiting}, or as an abstract structure graph (ASG) \cite{sievers2017structural}. Therefore, following the success of {\em Delfi}, a first attempt was made to use Graph Neural Networks to directly learn on the graphs and capture structural information \cite{Ma2020}. 

In our work, we continue the line of work on using GNNs directly for online planner selection.
We focus on exploring four GNN architectures, two graph representations, various dataset features and multiple prediction tasks. We also combine the advantages of GNNs and Extreme Gradient Boosting (XGBoost) \cite{chen2016xgboost}. 
Our main contributions are as follows: We explore the strengths and weaknesses of the GNN architectures for online planner selection. We investigate two graph representations, namely the lifted and grounded representation, in combination with additional node features. We focus on different ways to pick a planner including predicting the probability that a planner solves a task and the time it takes a planner for solving the task. Additionally, we experiment with the use of graph representations obtained by the GNNs for other ML-based methods like XGBoost.

\section{Related Work} 
In this Section, we embed our work into related approaches by highlighting and comparing them to ours. We provide an overview of recent ML-based approaches as well as approaches using Graph Neural Networks (GNNs). We incorporate related methods solving various tasks in the planning domain, including approaches using the IPC dataset for automatic planner selection.

Cenamor et al. \cite{cenamor2016ibacop} focus on creating a configurable portfolio which adapts to a given planning task. First, a number of candidate planners is selected with a filtering method. After that, predictive models estimate the time it takes to solve a task. Based on the combined results of the predictive models, it is decided which planners to include in the portfolio. Instead of creating portfolios, our work focuses on the task of directly choosing a planner for a given planning problem.

The Delfi planner \cite{katz2018delfi} makes use of graphical representations of a planning task in combination with deep learning techniques. Each planning problem of the IPC dataset is converted to an image and a CNN is used to make predictions. By applying it to new benchmarks, the authors show a good generalization of the proposed model. We also use the IPC data, but instead of CNNs, we experiment with GNNs. 

Sievers et al. \cite{Sievers2019} further pursue the method of converting planning problems to images and applying a CNN model. They analyze the shortcomings of the current methods and present possible solutions. Their main findings include the need for methods working well with non-IID data and exploring alternative network architectures. They especially analyze the shortcomings and possible solutions of existing CNN-based approaches. However, we mainly analyze how we can use GNNs to solve the automatic planner selection task. 

Other work using the IPC dataset is by Ferber and Seipp \cite{ferber2022explainable}. The focus of this work is to explore graph features which are used during training simple machine learning models, for instance, linear regression or random forests. The authors analyze the features and their importance for the training process. We distinguish ourselves by exploring different features when using different GNN models.

The closest to our work is the previous work on using GNNs for Automatic Planner Selection is by Ma et al. \cite{Ma2020}. The authors propose using two GNN architectures, namely GCN and GGNN, to select candidate planners. Their experiments with the IPC dataset show that their graph-based approach outperforms previous image-based ones. They highlight the ability of GNNs to capture structural information of a planning graph and address the lack of node-level information in previous approaches. The authors show that the lifted representation is favored over the grounded one as it produces more consistent results. However, it should be noted that the lifted representation contains much larger graphs and therefore scalable training approaches are needed. We distinguish ourselves by not only including two GNN architectures, but a set of four representative architectures with different focuses. In addition, we explore the use of node features for GNN training and combine GNNs with a simpler ML-based method. 

\section{Preliminaries}
The following covers important preliminaries our experiments are based on. This includes GNNs and its variants in Section~\ref{pre:gnn}, as well as XGBoost in Section~\ref{pre:boost}. 

\subsection{Graph Neural Networks} \label{pre:gnn}
Graph Neural Networks are used in numerous domains and capture the given graph structure. Node-level, edge-level or graph-level tasks can be solved. The training incorporates two main steps, namely \texttt{aggregate} and \texttt{update}. First, the node representations of all neighboring nodes are aggregated according to
\begin{equation}
    a_{v}^{(t+1)} = AGGREGATE^{(t+1)}({h_{u}^{(t)}}: u \in N_{(v)})
\end{equation} with the node representations at the $t$-th layer $h_{u}^{t}$ and the set of neighbors of target node $v$ $N_v$. Thereafter, the node representations are combined and the target node is updated using
\begin{equation}
    h_{(v)}^{(t+1)}=UPDATE^{(t+1)}(h_{v}^{(t)}, a_{v}^{(t+1)})
\end{equation}
The main difference between different GNN architectures is the choice of $AGGREGATE^{(t+1)}(.)$ and $UPDATE^{(t+1)}(.)$  \cite{hamilton2020graph}. Various methods have been developed, we will cover four of them in the following.

\subsubsection{Graph Convolutional Network (GCN)}
One commonly used GNN architecture is the Graph Convolutional Network (GCN) \cite{kipf2016semi}. Inspired by convolutions used for images, GCN use convolution filters that operate directly on the graph structure. In contrast to images, the neighborhood size of a node within a graph varies. Therefore, a parameter matrix transforms the node representations obtained from the previous layer. The transformed representations are weighted according to the graph adjacency matrix \cite{kipf2016semi,Ma2020}. When using GCN, an update step is defined as 
\begin{equation}
    H^{(t+1)} = \sigma(\hat{A}H^{(t)}W^{(t)})
\end{equation}
where $H^{(t+1)}$ denotes the matrix with stacked node representations $h_{v}^{t}$, $v$ is a node and $t$ stands for the current layer. $\sigma$ is an activation function (e.g., ReLU), the adjacency matrix $A$ is normalized to $\hat{A}$ and $W$ is the parameter matrix. GCN uses a shared weight for all edges, making the model relatively simple. In case of more complex graph structures, this approach might be less expressive.

\subsubsection{Gated Graph Neural Network (GGNN)}
Gated Graph Neural Networks (GGNNs) \cite{li2016gated} distinguish themselves from other architectures by incorporating gated recurrent units (GRUs) \cite{cho2014learning} in their propagation module. The current state of the nodes are updated by the GRU which views the nodes and their representations as a dynamic system. The node representations are updated as follows:
\begin{equation}
    h_{v}^{(t+1)}=GRU(h_{v}^{(t)}, m_{v}^{(t+1)})
\end{equation}
Here, $m_{v}^{(t+1)}$ is a message which is aggregated in order to update the state of the node. This formulation allows for deploying selective updates of the node representations depending on the information aggregated from the neighbors. Besides local information, long-range dependencies within the graph can be captured. 

\subsubsection{Graph Attention Network (GAT)} \label{sec:gat}
Instead of graph convolutions, Graph Attention Networks (GATs) \cite{velivckovic2018graph} use masked self-attentional layers. Different weights are assigned to different neighbors when aggregating the neighboring features. This enables the nodes to focus on the more relevant neighbors and helps the model to capture complex relationships in the graph. Another advantage of this approach is that the importance of the neighboring nodes is determined without knowing the graph structure beforehand. The update node representations can be obtained with 
\begin{equation}
    h_{v}^{(t+1)} = \sigma (\frac{1}{K} \sum_{k=1}^{K} \sum_{u \in N_v} \alpha_{vu}^{k} W_k h_{u}^{t})
\end{equation}
with the normalized attention coefficient $\alpha_{vu}^{k}$ for the $k$-th attention head and the weight matrix $W$. By employing attention, the model is particularly effective for detecting local, but also global dependencies.

\subsubsection{Graph Isomporhism Network (GIN)}
Another variant is the Graph Isomorphism Network (GIN) \cite{xu2018powerful} which makes use of multi-layer perceptrons (MLPs) to learn the parameters of the update function. It is inspired by the Weisfeiler-Lehman (WL) graph isomporhism test \cite{leman1968reduction} which determines how \textit{similar} two graphs are. A node update is defined by 
\begin{equation}
    h_{v}^{(t+1)} = MLP^{(t+1)} ((1+ e^{(t+1)}) * h_{v}^{(t)} + \sum_{u \in N(v)}h_{u}^{(t)})
\end{equation}
where $\varepsilon $ is a learnable parameter. GIN uses a simple \textit{sum} operator to aggregate the features which makes it computationally efficient. Through the learnable parameter, it is able to adapt well to various graph structures and can effectively capture graph-level features making it a common choice for solving graph-level tasks.

\subsection{Extreme Gradient Boosting} \label{pre:boost}
XGBoost, short for Extreme Gradient Boosting, is a machine learning technique combining multiple decision trees to create a strong model \cite{chen2016xgboost}. The decision trees often have limited depth and the predictions of each tree are added up to obtain the final prediction. With the help of an objective function, gradient optimization is performed and a learner is fitted with respect to the current predictions. The objective function is minimized during training and comprises the loss function and a regularization term. A series of decision trees is built sequentially, with each tree correcting the errors of the combined model up to that point. The objective function at the $t$-th iteration is given through 
\begin{equation}
    \mathcal{L}^{(t)} = \sum_{(i=1)}^{n} l(y_i, \hat{y}_{i}^{(t-1)} + f_t(x_i)) + \Omega(f_t) 
\end{equation}
where the loss is calculated according to a loss function $l$ based on the prediction $\hat{y_i}$ and the target $y_i$ at the $i$-th instance. The tree structure $f_t$ which gains the most improvement of the model is added. The regulariztaion term $\Omega(f_t)$ ensures the model is kept as simple as possible. 
In general, XGBoost combines the strengths of decision trees with an optimization process to efficiently and effectively handle complex datasets. 

\section{Experiments}
In this section, we describe the methodology of our experiments including the datasets, configurations, tasks and setup. This is followed by the presentation and analysis of results and their implications. Lastly, we summarize the main findings and insights.

\subsection{Methodology}

\subsubsection{Dataset and features} 
\begin{figure}[t]
    \centering
    \includegraphics[width=0.52\textwidth]{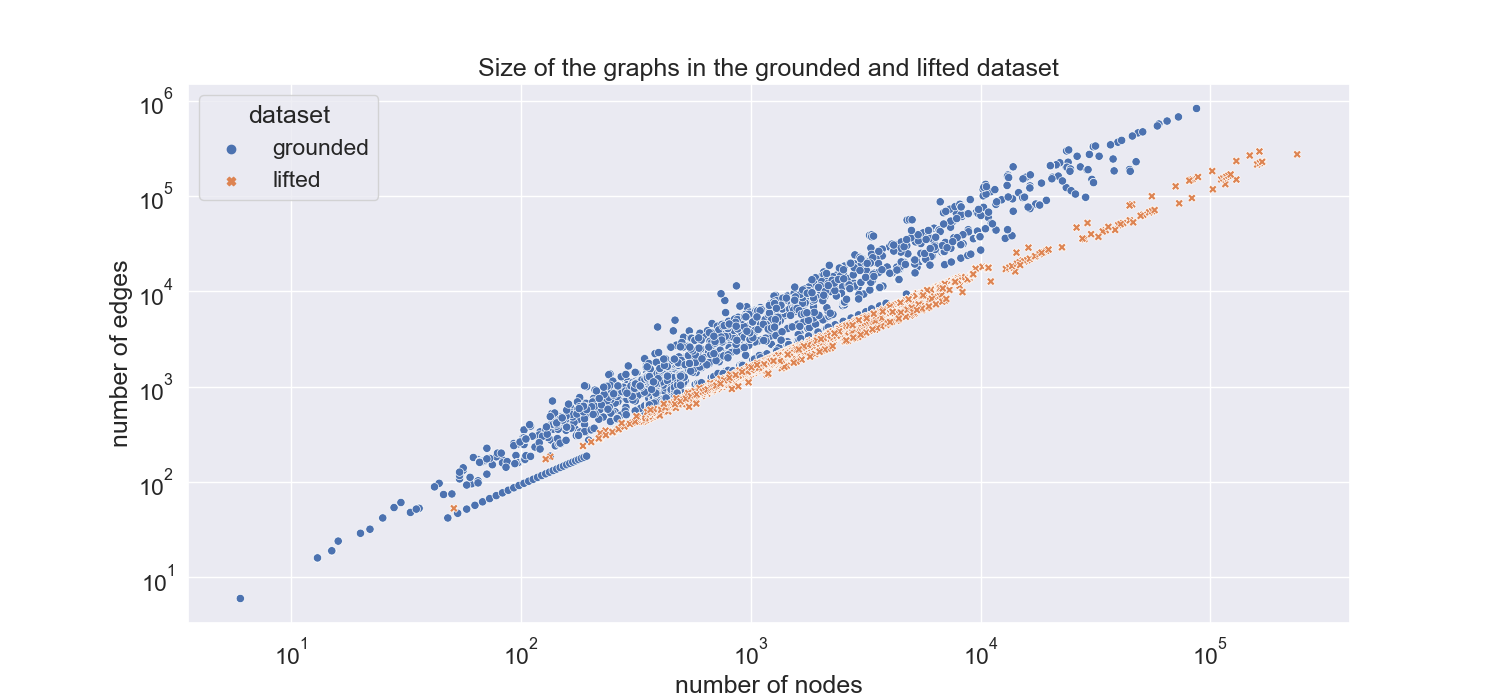}
    \caption{Graph sizes of the grounded and lifted representations}
    \label{fig:graphsizes}
\end{figure}

We use the publicly available dataset consisting of tasks from various
International Planning Competitions (IPC), with their graph representations and planners performance data \cite{Ferber2019}. It consists of 2439 data points and provides splits for 10-fold cross validation (2294/145 training+validation/testing). The split can either be random or domain-preserving, making sure that planning problems of the same domain are not split. In the portfolio, there are 17 planners. For each planner, the target value is the time needed for solving the problem. In case a planner exceeds the timeout limit of 1800 seconds, the target value is set to 10,000. 
There are two representations, the grounded and the lifted representation. While the grounded one is based on SAS+ \cite{backstrom1995complexity} and directed Problem Description Graphs (PDG) \cite{pochter2011exploiting}, the lifted representation is based on the Planning Domain Definition Language (PDDL) \cite{mcdermott20001998} and directed acyclic Abstract Structure Graphs (ASG) \cite{sievers2017structural}. Figure~\ref{fig:graphsizes} illustrates the number of nodes and edges for each graph in the grounded and lifted representation. The grounded graphs consist of up to 100,000 nodes and 800,000 edges. The lifted graphs contain up to 250,000 nodes and 300,000 edges, meaning generally more nodes, but less edges than the grounded representation. This is also reflected by the average node degree which is 12.26 for the grounded graphs and 2.92 for the lifted ones. 

\begin{figure}[t]
    \centering
\includegraphics[width=1.1\columnwidth]{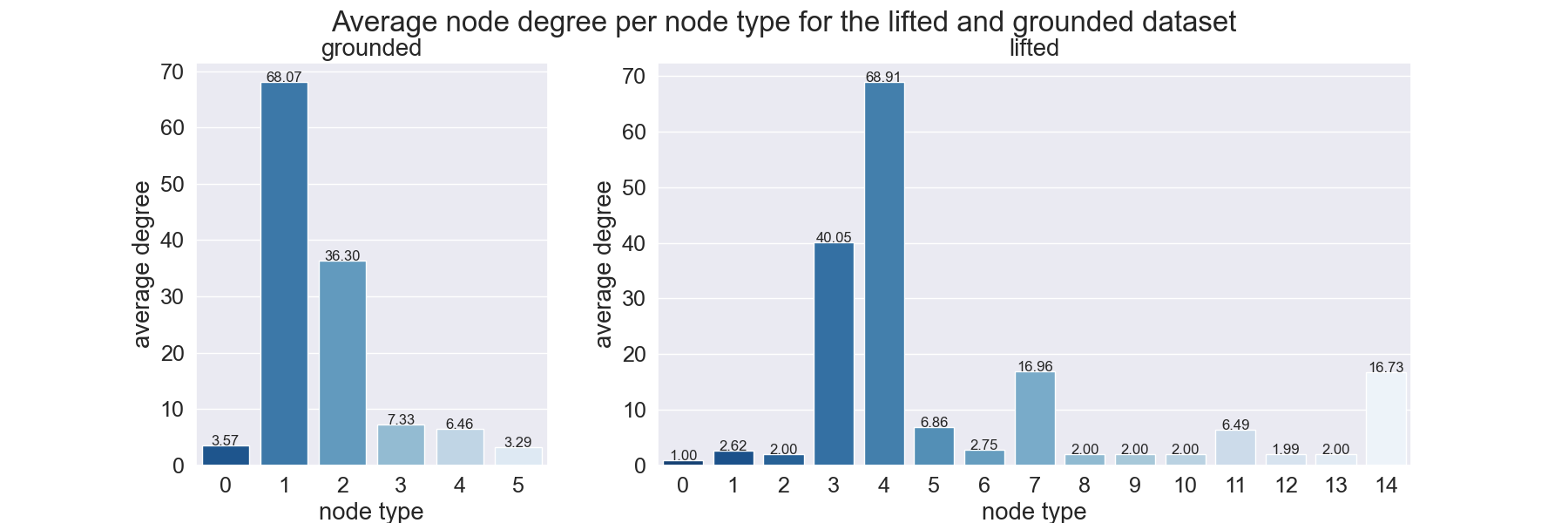}
    \caption{Average node degree per node type for the grounded dataset.}
    \label{fig:nodedegree}
\end{figure}

As node features, the node type is used and encoded into one-hot vectors. For the lifted representation, there are 15 different node types and for the grounded representation, there are 6 individual node types. For instance, the type can depict that the node represents a \texttt{constant}, an \texttt{action}, or an \texttt{effect}. We analyse the average node degree per node type, depicted in Figure \ref{fig:nodedegree}. For both the grounded and the lifted graphs, while there are high-degree node types with an average degree of up to 70, the majority of node types have an average degree of 3 to 8 for the grounded graphs and 1 to 17 for the lifted ones. 

Due to the difference of average node degree per node type, we decide to enhance the initial node features with the node degree. In addition, the type of the neighboring nodes of a target node is important within planning problems to get a more detailed context of the node within the graph. Therefore, we also experiment with the type of the neighbors as node feature.

\subsubsection{Tasks and configurations}
We investigate multiple methods to select a planner.
The first method is to predict the time to solve the planning problem and choose the planner with the best predicted time performance, henceforth denoted by {\em time}.  
The second is to predict how likely it is for a planner to solve the planning problem within the overall time bound of 1800s and choose the planner with the highest probability to solve within the bound, henceforth denoted by {\em binary}. 
These two methods were explored in previous work with CNNs \cite{Sievers2019}. 
Four GNN architectures are used to obtain the predictions, namely GCN \cite{kipf2016semi}, GGNN \cite{li2016gated}, GAT \cite{velivckovic2018graph} and GIN \cite{xu2018powerful}. We mainly explore two additional node features: the node degree and the type of the neighboring nodes. 
The third method has not been explored so far in the context of online planner selection. We use the graph representation obtained by the last layer of a GNN as input to train a classification task with a XGBoost model \cite{chen2016xgboost}.

\subsubsection{Training details} 
We choose the Deep Graph Library (DGL) \cite{wang2019dgl} based on PyTorch \cite{paszke2019pytorch} as a framework. The configurations are oriented on \cite{Ma2020}. For training, we use the Adam optimizer \cite{KingBa15} with learning rate 0.001. The number of layers is set to 2, the size of the hidden dimension is 100 and we train the model for 100 epochs. For regression tasks, the MSE loss is used and for the other tasks, we choose the binary cross entropy loss. We run the experiments either on a Nvidia P100 GPU or a Nvidia RTX A600 with two Intel(R) Xeon(R) CPU E5-2640 v4 or two AMD EPYC 7282 CPU nodes, respectively. Depending on the model, training takes 10 to 60 minutes. We set the number of estimators to 500, the maximum depth to 5, and the learning rate to 0.01 when using XGBoost. Training is done with early stopping after 20 epochs.

\subsection{Results}
First, we present and analyze the results for all four GNN architectures, the two tasks based on probability and time, and the lifted and grounded representation in combination with the random and domain-preserving split. Subsequently, we explore how the results change when adding node features to the data. We conclude by illustrating the accuracy changes when combining a GNN model with XGBoost. 
It is important to note that we repeat each experiment 10 times and present the corresponding average accuracy with the standard deviation. 

\begin{figure}[t]
    \centering
    \includegraphics[width=\columnwidth]{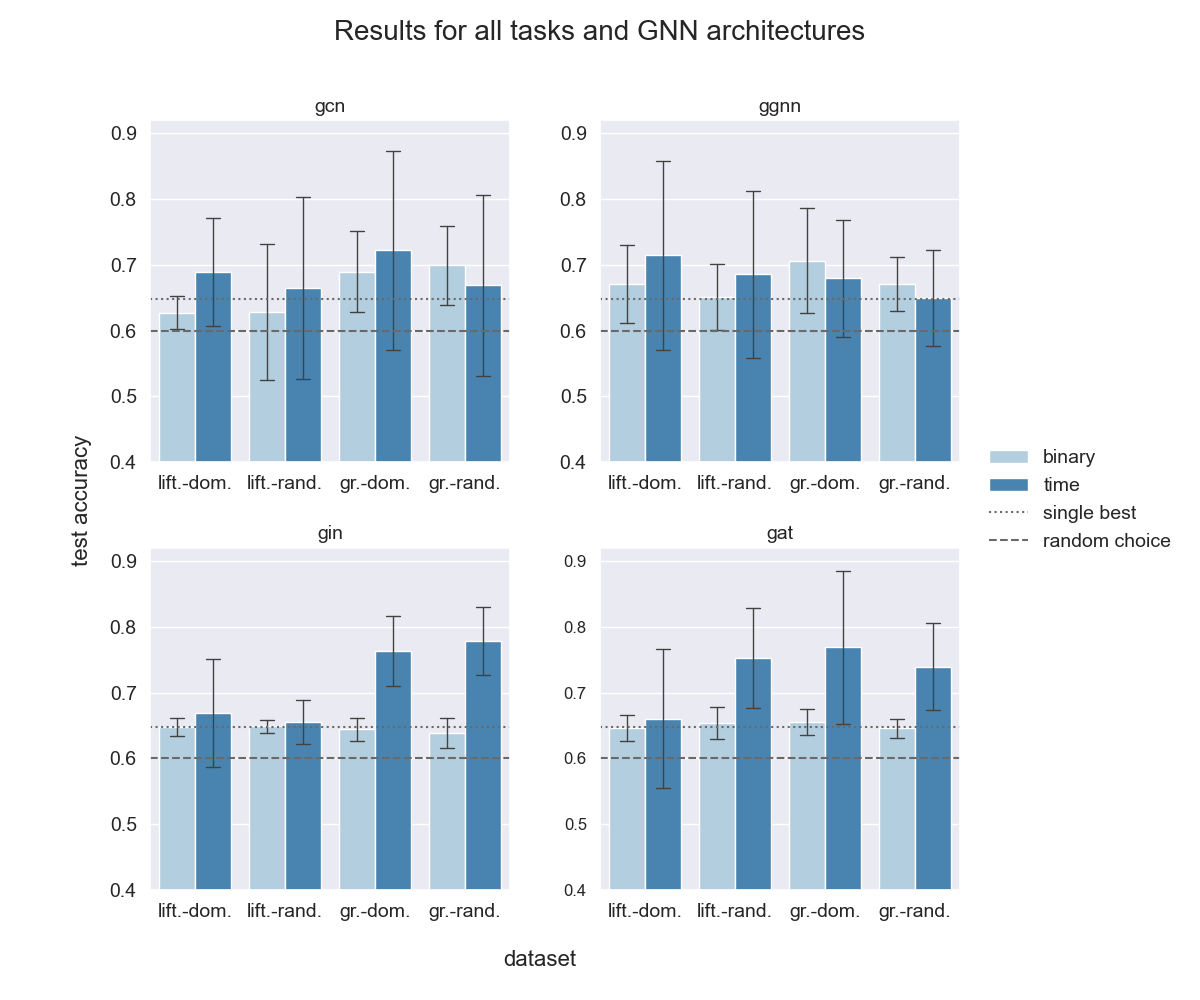}
    \caption{Results for all tasks and all GNN architectures}
    \label{fig:resall}
\end{figure}
\subsubsection{Base experiments} \label{sec:baseexp}
\begin{figure}
    \centering
    \begin{subfigure}{0.48\columnwidth}
        \includegraphics[width=\textwidth]{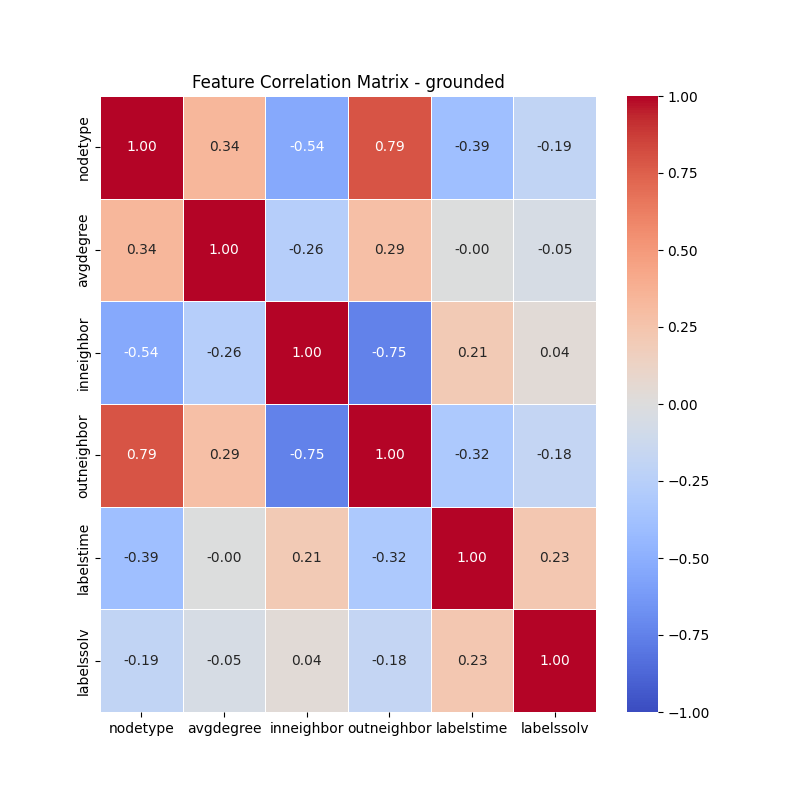}
        \caption{grounded}
        \label{fig:corrgrounded}
    \end{subfigure}
    \begin{subfigure}{0.48\columnwidth}
        \includegraphics[width=\textwidth]{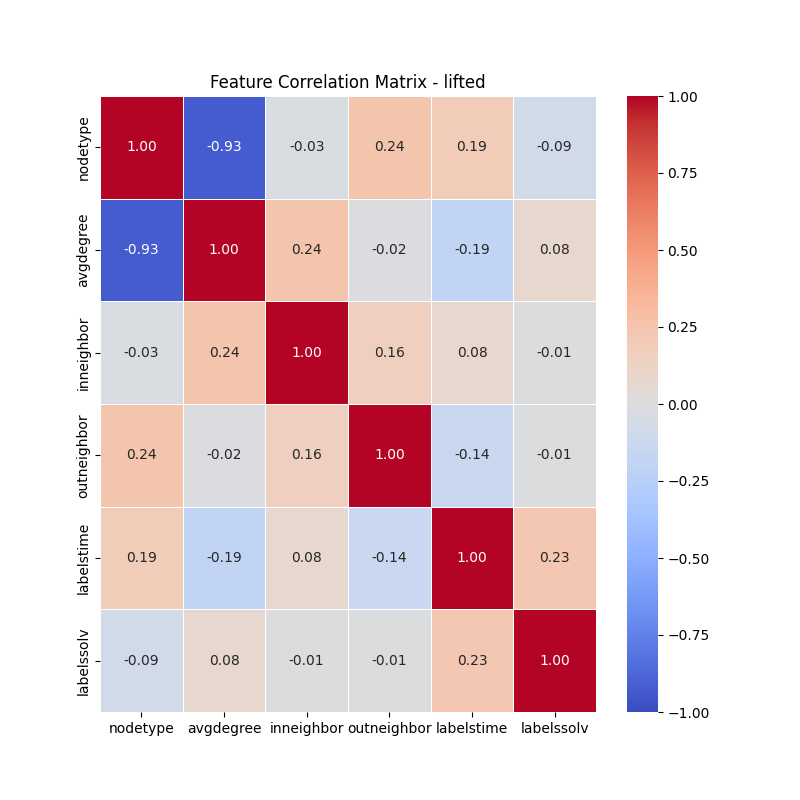}
        \caption{lifted}
        \label{fig:corrlifted}
    \end{subfigure}
    \caption{Feature correlation matrix for the time- and solvable-based task for the grounded and lifted dataset}
    \label{fig:corr}
\end{figure}

This experiment aims to investigate how different GNN architectures perform combined with the two graph representations and to find out how well the planning problem is captured by them. Figure \ref{fig:resall} gives an overview of the results. For all four GNN architectures, we plot the accuracy for the lifted and grounded dataset as well as the random and domain-preserving split. Overall, the task where we predict the time performs better than the one predicting whether a planning problem is solvable. Figure \ref{fig:corrgrounded} shows the correlation of the node type to the time- and solvable-based labels. For the grounded dataset, we can see a higher negative correlation between node type and the time-based labels than the solvable-based ones with values of -0.39 and -0.19, respectively. Analogous to the grounded dataset, the correlation matrix for the lifted dataset (Fig. \ref{fig:corrlifted}) shows that the time-based labels also have a higher correlation than the solvable-based labels which explains the better performance when predicting the time. 
For almost all GNN architectures, the grounded representation obtains better results than the lifted one, especially together with the domain-preserving split. This is also reflected in the correlation matrices where the correlation between the node type and the labels is higher for the grounded graphs than the lifted ones (see Fig. \ref{fig:corr}). 
When looking at the results, we see that the standard deviation is quite high, especially for the experiments where the time is predicted. Although the best performance can be obtained when predicting the time in most cases, it is also not as stable as predicting a probability. This lies in the nature of the problem. Predicting the actual runtime of a planner is a very difficult task, but it is also a task working very well. To reduce variance, one could incorporate the top-k predicted planners and then refine the predictions for those by using simple ML methods. 

\begin{figure}[t]
    \centering
    \includegraphics[width=\columnwidth]{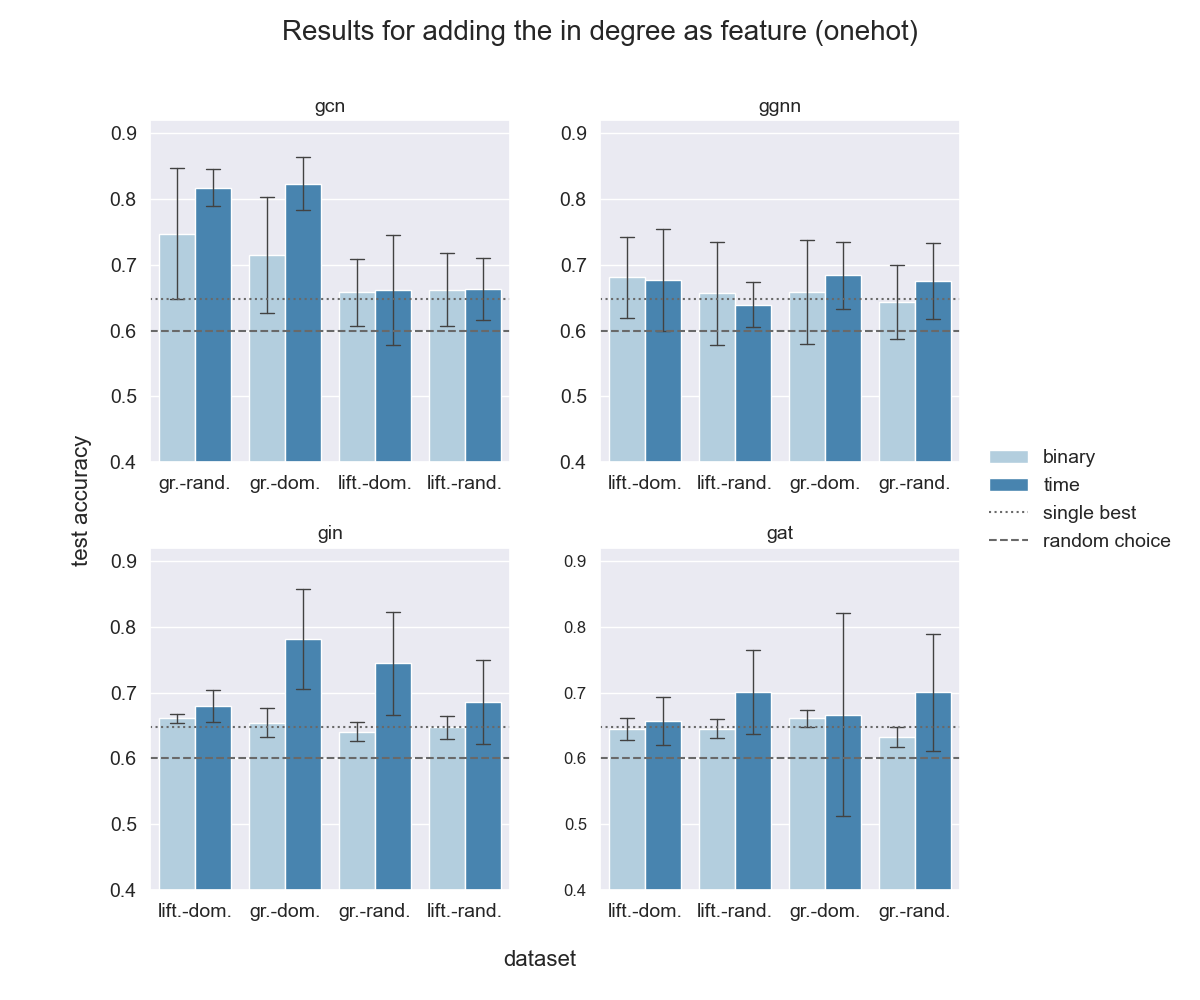}
    \caption{Results for adding the in degree as a node feature}
    \label{fig:indegree}
\end{figure}

In general, GCN and GGNN show a similar performance with an accuracy slightly over 0.7 for the time-based task. GIN and GAT obtain an even higher accuracy around 0.77, again for the time-based task. GIN is based on the WL graph isomorphism test (Section \ref{sec:gat}) to check how similar two graphs are. GIN has been proven to be especially effective for graph-level tasks which is supported by our experiments. For planning problems, the direct predecessors and successors of a node are relevant as they indicate how an action, effect, or axiom are embedded in the overall problem. GAT emphasizes more important nodes by applying its attention mechanism, while GGNN tries to incorporate far away neighbors. The results illustrate that GGNN is not as effective as GAT at capturing a planning problem. A reason can be that far away neighbors are not as relevant as direct neighbors for the graph-level predictions. To obtain the graph-level predictions, we apply a pooling layer. By including far away neighbors on the node-level, no additional information is gained as the information is pooled in the end anyways. 

\begin{figure}[t]
    \centering
    \includegraphics[width=\columnwidth]{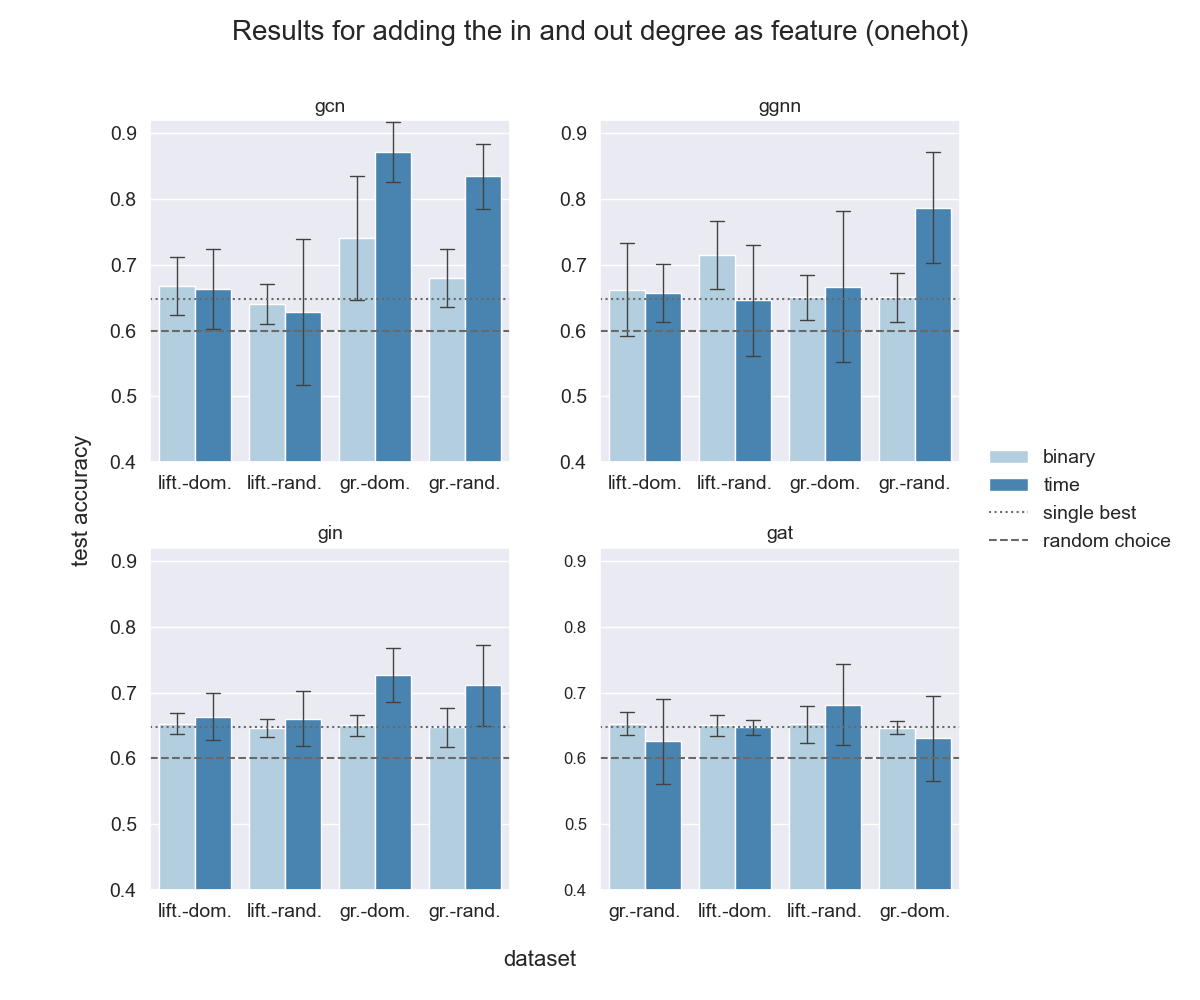}
    \caption{Results for adding the in and out degree as a node feature}
    \label{fig:inoutdegree}
\end{figure}
\subsubsection{Enhancing the node features}
In the following, we enhance the dataset with hand-picked features and investigate how they influence the results. Before diving into the experiments, we investigate the feature correlation. Figure \ref{fig:corr} gives an overview of the correlation of node type, average degree, incoming and outgoing neighbor type with the target labels (either time- or solvable-based). We can see that the average degree has a correlation of -0.19 for the lifted dataset, however nearly no correlation is detected in the grounded dataset. The type of the neighboring nodes has a up to -0.32 for the grounded dataset, whereas the correlation in the lifted datset is relatively low. A reason for these differences are the different represeantions and their characteristics. Another reason could be that we correlate the average node features of the graphs with the output labels instead of the individual node features, to be able to use numeric values for the calculations. Thus, this might not be completely representative, but gives a direction of the correlation. To verify this theory, we need to look at the experimental results. 

First, we add the node degree as node feature. We investigate adding the in-degree, the out-degree and both. Within a planning problem which is modeled as a directed graph, the subsequent transitions are explicitly known through the formal description of the planning problem. The preceding transitions, on the other hand, are only implicitly known. Therefore, we suspect an improvement of the results especially when incorporating the in-degree of each node. We model the feature as a one-hot encoded vector and append it to the vector describing the node type. As shown in Figure \ref{fig:indegree}, including the in-degree improves the results to an accuracy up to 0.81 when using the GCN with the grounded dataset. It is interesting to see that the accuracy with GAT slightly decreases compared to using only the node type as a feature (Fig. \ref{fig:resall}). As GAT already captures local information very well, redundant information is added when using the node degree as feature leading to a performance decrease. GGNN and GIN do not show any remarkable changes. 

Looking at the results with in- and out-degree as node feature (see Fig. \ref{fig:inoutdegree}), the most outstanding result is the increase of accuracy for the GCN rising up to 0.87 with the grounded dataset. GGNN also shows some improvement up to an accuracy of 0.8 (grounded dataset). Local node characteristics can influence surrounding and far away nodes, as in a planning problem, a transition not only influences the following steps, but also further steps in the future. Therefore, when using GGNN, the node degree as feature enhances the training as it emphasizes sequences and not only local neighborhoods. 

\begin{figure}[t]
    \centering
    \includegraphics[width=\columnwidth]{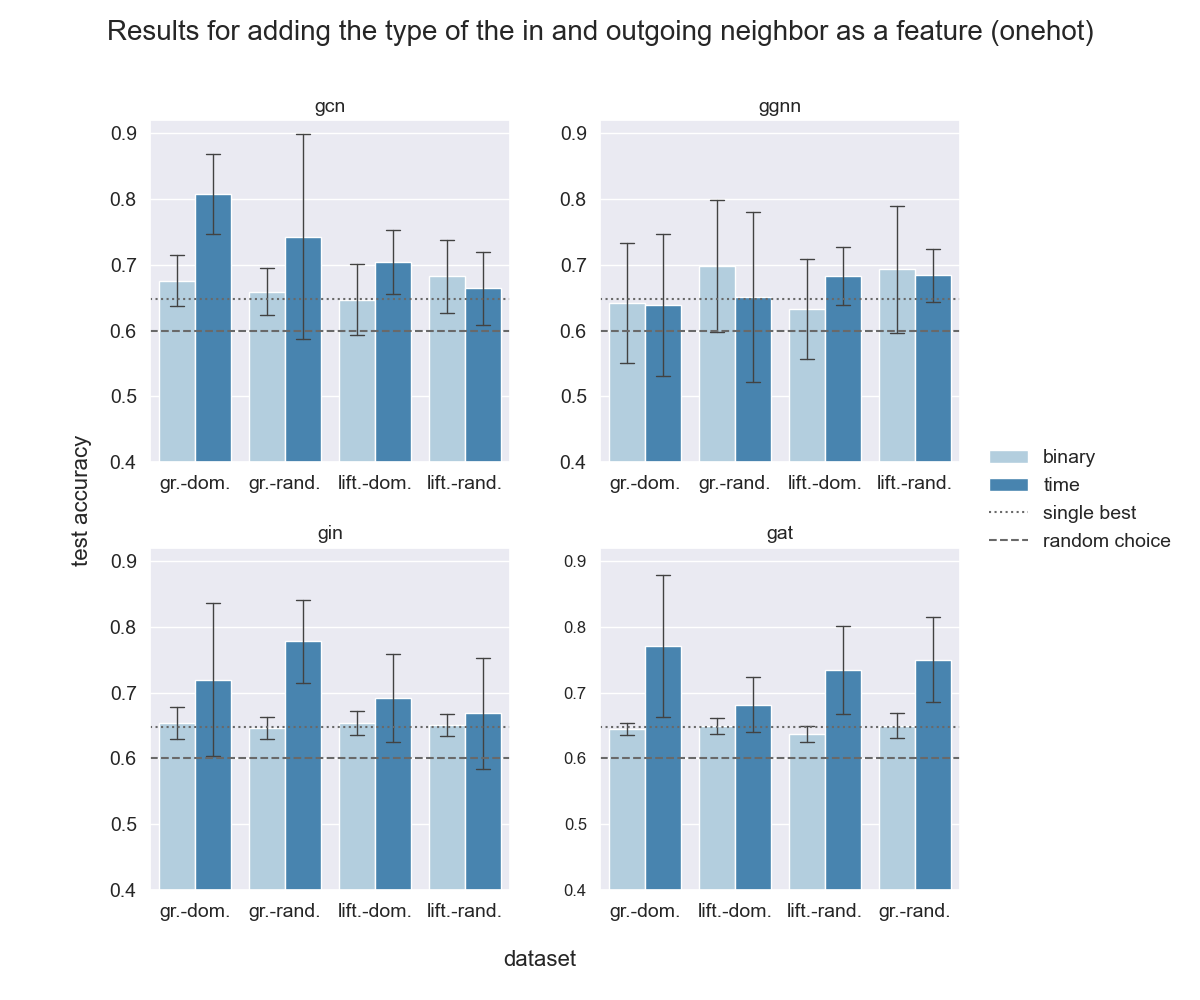}
    \caption{Results for adding the in- and out-going neighbor type}
    \label{fig:inoutneighbors}
\end{figure}

For each step in a planning task, it is important to know what type of node follows or precedes another one. Therefore, we include the type of the neighboring nodes as node feature. In Fig. \ref{fig:inoutneighbors}, we see can see a higher accuracy for GCN when using the node type only. Although there are some changes compared to the results of the basic experiments, the improvements are not as high as with adding the node degree. This can be explained by how GNNs work in general. In each epoch, each node passes its current information to its neighbors. Among other things, the node type is transmitted. Thus, the neighboring node type is captured anyways, without needing to add it as a feature. By adding it, we emphasize the importance of the node type, but do not necessary gain new information or insights for model training. Further, when looking at the feature correlation matrix (Fig. \ref{fig:corr}), the correlation of the neighbor type especially for the lifted dataset is much lower compared to the node degree. 

\begin{figure}[t]
    \centering
    \includegraphics[width=\columnwidth]{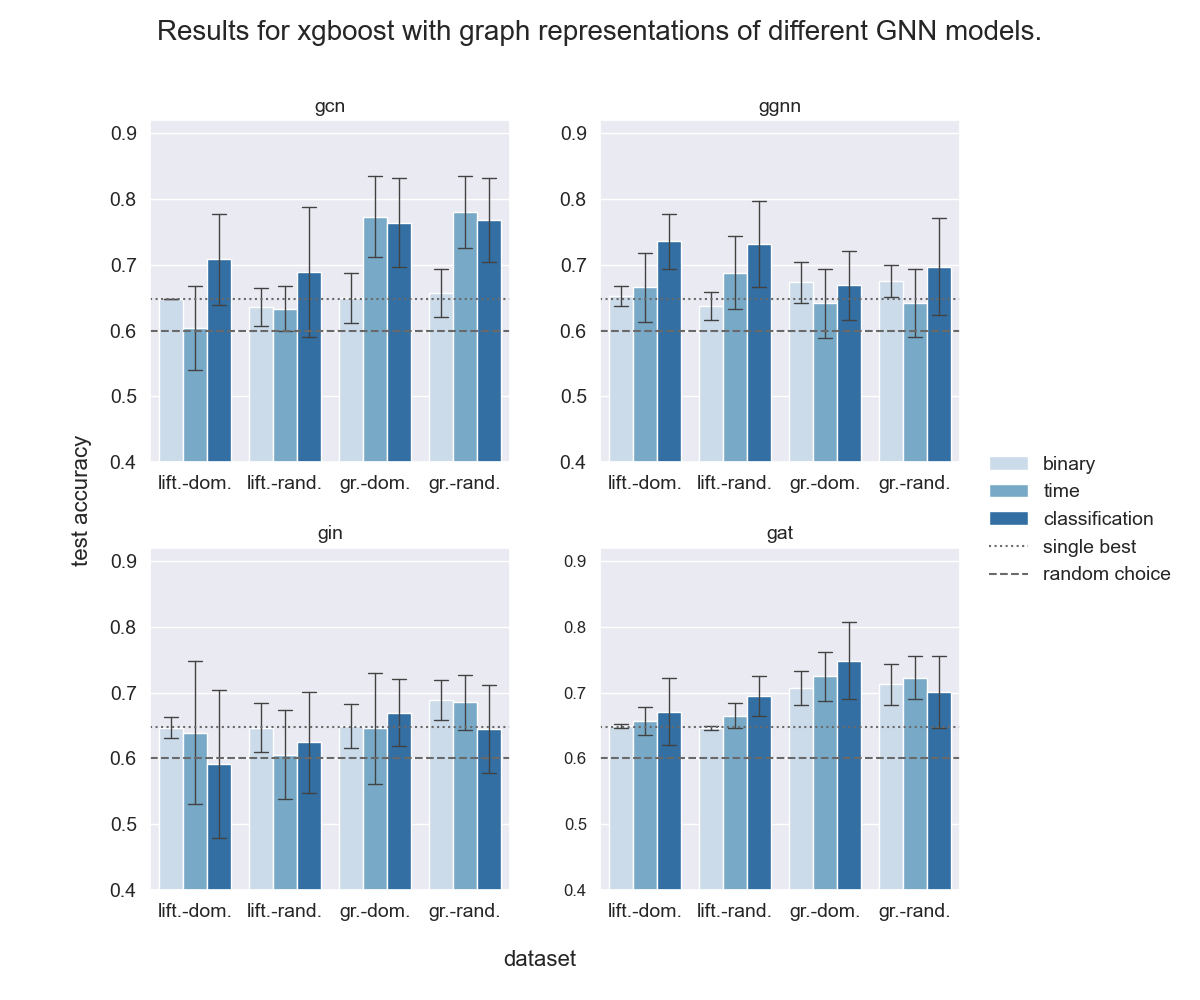}
    \caption{Results for combining GNNs with XGBoost}
    \label{fig:xgboost}
\end{figure}
\subsubsection{Combining GNNs with XGBoost}  
Instead of training a single GNN model, we combine the strengths of GNNs and gradient boosting in the following experiments. We use the GNN to obtain a graph representation as one strength of GNNs is the ability to capture graph structures. This representation is 100-dimensional and is used as input for training an expressive XGBoost model. We train XGBoost models in the following three ways: predict the probability and then pick the planner with the highest probability, predict the time and pick the planner which uses the least amount of time, directly pick one of the 17 planners through multi-class classification. We use the default features provided by the dataset, namely the node type. 

In Fig. \ref{fig:xgboost}, the accuracy for the experiments with all four GNN architectures and all three tasks is shown. GCN and GAT obtain the overall best results, especially for the time-based and classification-based task with values of slightly under 0.8. Again, the results with the grounded dataset are better than the lifted one which is caused by the higher correlation between the features (e.g., node type) and the labels. When using GCN, predicting the time produces the best results, but the classification task is comparable. In case of GAT, the classification works best. Here, the grounded dataset and the domain-preserving split are used. Using a GNN for the classification task could not produce satisfactory results with accuracies of under 0.5. However, the combination of GNNs and XGBoost shows a much better performance and is similar to the results with GNNs only and the time-based task (Fig. \ref{fig:resall}). Another benefit of combining GNNs with XGBoost is the resource-utilization. While we have to train the GNN-only experiments on GPUs, we don't necessarily need GPU-training for XGBoost. Training the GNN plus XGBoost method on our CPU takes a similar amount of time on the CPU as the GNN-only experiments on our GPU.

\section{Discussion and Conclusions}

\subsection{Discussion \& Insights}
We have seen that GNNs are able to effectively capture planning problems. The best performing way to pick a planner is by predicting the time and then choosing the best planner based on the predictions. Without adding additional features, GIN and GAT prove to work well with an accuracy of up to 0.87. It is shown that by adding simple graph features like the node degree, the accuracy can be improved up to 0.9. The type of the neighbor is also important, but introduces redundant information for some models not reaching the performance obtained with the node degree. 

GNNs combined with XGBoost are good at directly picking a planner without having to predict things like the time. Although the accuracy is only comparable to our basic GNN-only experiments, training XGBoost can be done on CPUs instead of GPUs in a similiar amount of time. Thus, this method produces good results and proves to be more resource efficient. Further, multiple GNNs can be used to add different graph representations which helps to capture a planning problem of different perspectives. The results indicate that there is potential, but still room for improvement. 

\subsection{Future Work}
This section gives an overview of future research directions based on our findings. First, we use quite simple graph-based features for training like the node type and the node degree. To further improve the results, one could include more features like the node centrality, information about clusters or learned node embeddings with tools like node2vec. 

Another possible direction could be to use a mixture of experts model \cite{shazeer2017outrageously} or to train a separate GNN for each of the 17 target planners to make them more specialized and better at deciding whether a planner solves a task or not. Alternatively, the process of selecting a planner could be divided into two phases: first, the top-k planners are chosen and then, the selection is refined by focusing only on the top-k planners instead of all 17. 

We use standard GNN architectures for our experiments which work on a variety of tasks. However, they are not adapted to automatic planner selection. Important is the ability to directly select a planner without the need to predict other features like the time first. Ideas of the well-working GNNs in our experiments can be combined with knowledge of how planning problems are formally described and transformed to graphs. For instance, most planning problems are directed acyclic graphs. This knowledge could be a large benefit for the resulting specialized architectures.

When training the GNN models for XGBoost, one could use the enhanced datasets with richer node features. This could help getting a more precise graph representation which would then increase the results of XGBoost. Instead, graph-level features could be included leading to a more concrete description of the graph. Possible graph features are the size of the graph, the clustering coefficient, the majority node type, the diameter or the strongly connected components.

\subsection{Conclusions}
Automatic planner selection can be tackled using different ML-based approaches, including GNNs. In our work, we investigated the use of different GNNs for choosing a planner for a given planning problem. We explore four GNN architectures, two graph representations, various node features and different ways to pick a planner. Overall, the grounded dataset showed a better performance than the lifted one which is due to the higher feature correlation. Further, predicting the time and then picking a planner based on that led to an accuracy of up to 0.87 when using a GCN model. We analyse the characteristics of the four models to understand what is important for improving automatic planner selection. Further, we investigate the influence of different node features. It is shown that by adding the node degree, the results can easily be improved. Combining GNNs with XGBoost allows to train a classification task where a planner is directly chosen instead of first predicting the time and then make the decision based on that. Our results show a similar accuracy to the ones obtained in our basic GNN-only experiments, but there is no need for GPUs in contrast to GNN training. In addition, we combine multiple graph representations obtained from multiple GNNs to improve the results of XGBoost. To conclude, automatic planner selection with GNNs shows a good performance and can be done in different ways. We obtain the best results with an accuracy of up to 0.9 with the GCN, the grounded representation and the node degree as a feature.

\appendix

\bibliographystyle{named}
\bibliography{bib.bib}

\end{document}